\pdfoutput=1

\documentclass[11pt]{article}

\usepackage{coling}
\usepackage{tabularx}
\usepackage{booktabs}
\usepackage{times}
\usepackage{latexsym}
\usepackage{bbding}
\usepackage{amsmath} 
\usepackage{amsfonts}
\usepackage{algorithm}
\usepackage{algpseudocode}
\usepackage{algorithmicx}
\usepackage{threeparttable}
\usepackage{amsmath} 
\usepackage{mdframed}
\usepackage[most]{tcolorbox}
\usepackage{xcolor}
\usepackage{multirow}
\usepackage{CJKutf8}

\usepackage[T1]{fontenc}

\usepackage[utf8]{inputenc}

\usepackage{microtype}

\usepackage{inconsolata}

\usepackage{graphicx}
\usepackage{color}

\title{Improvement in Sign Language Translation Using Text CTC Alignment}

\author{
  \textbf{Sihan Tan\textsuperscript{1,2}},
  \textbf{Taro Miyazaki\textsuperscript{2}},
  \textbf{Nabeela Khan\textsuperscript{1}},
  \textbf{Kazuhiro Nakadai\textsuperscript{1}}
\\
  \textsuperscript{1}Institute of Science Tokyo,
  \textsuperscript{2}NHK Science \& Technology Research Laboratories
\\
\texttt{\{tansihan, nabeela, nakadai\} \href{mailto:tansihan@ra.sc.e.titech.ac.jp}{@ra.sc.e.titech.ac.jp}}\\ \href{mailto:miyazaki.t-jw@nhk.or.jp}{\texttt{miyazaki.t-jw@nhk.or.jp}}
  }

\begin{document}
\maketitle
\begin{abstract}
Current sign language translation (SLT) approaches often rely on gloss-based supervision with Connectionist Temporal Classification (CTC), limiting their ability to handle non-monotonic alignments between sign language video and spoken text. In this work, we propose a novel method combining \textit{Joint CTC/Attention} and \textit{transfer learning}. The \textit{Joint CTC/Attention} introduces hierarchical encoding and integrates CTC with the attention mechanism during decoding, effectively managing both monotonic and non-monotonic alignments. Meanwhile, \textit{transfer learning} helps bridge the modality gap between vision and language in SLT. Experimental results on two widely adopted benchmarks, RWTH-PHOENIX-Weather 2014 T and CSL-Daily, show that our method achieves results comparable to state-of-the-art and outperforms the pure-attention baseline. Additionally, this work opens a new door for future research into \textit{gloss-free} SLT using text-based CTC alignment.\footnote{Code and resource are available: \url{https://github.com/Claire874/TextCTC-SLT}}

\end{abstract}

\section{Introduction}
\begin{figure}[t]
  \includegraphics[width=\columnwidth]{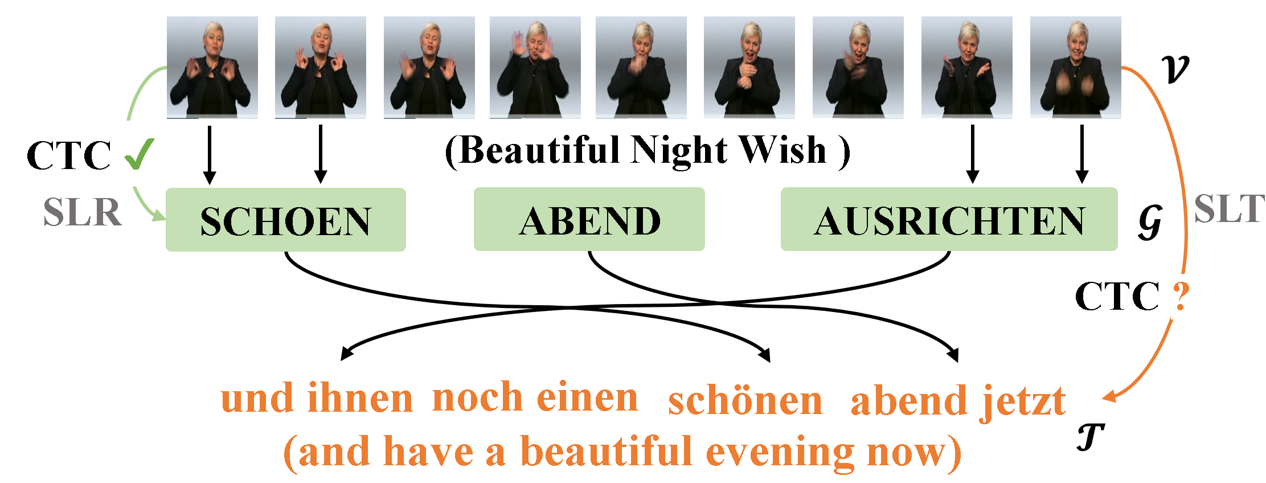}
  \centering
  \caption{An overview of sign language recognition and sign language translation alignment. CTC is used for monotonic alignment between the sign language video $\mathcal{V}$ and a gloss sequence $\mathcal{G}$. While the alignment between the sign language video with corresponding spoken text $\mathcal{T}$ is non-monotonic. }
  \label{fig:introduction}
\end{figure}

Sign language is a natural language used by the deaf and hard-of-hearing community for communication~\cite{Sutton-Spence_Woll_1999}, expressed through manual and non-manual elements, such as hand movements, fingerspelling, facial expressions, and body posture~\cite{introduction}. Due to the multi-modal nature of sign language, recent studies in sign language processing (SLP) aim to bridge the modality gap between vision and language. SLP encompasses several tasks, including sign language recognition (SLR)~\cite{zuo2023natural,sandoval2023self}, translation (SLT)~\cite{camgoz2018neural,zhang2023sltunet,gong2024llms}, and production~\cite{rastgoo2021sign}. In this paper, we focus on SLR and SLT. SLR involves identifying signs as glosses, a written representation of signs in the original order, whereas SLT focuses on translating the sign language video into its corresponding spoken text. Many SLR approaches utilize Connectionist Temporal Classification (CTC) loss~\cite{ctc} to localize and recognize glosses in an unsegmented sign language video~\cite{cui2017recurrent,min2021visual}. However, in the context of SLT, CTC loss is typically employed only when glosses serve as an intermediate or auxiliary form of supervision~\cite{camgoz2020sign, yin2020better,chen2022simple}. This reflects the assumption that CTC’s monotonic alignment between input and output sequences is suitable for gloss-based supervision~\cite{wong2024sign2gpt} but inadequate for the non-monotonic alignment required to directly translate sign language into corresponding spoken text, as displayed in Figure~\ref{fig:introduction}.

Nevertheless, many studies have successfully applied CTC to machine translation (MT)~\cite{libovicky-helcl-2018-end,saharia-etal-2020-non} and speech translation (ST) ~\cite{chuang-etal-2021-investigating,peng-etal-2024-owsm}, demonstrating that CTC’s reordering capability holds promise for handling complex translation tasks. These approaches show that while CTC is effective for monotonic alignment tasks, it can be extended to non-monotonic alignment to address reordering challenges. This leads us to the question: \textit{
Can CTC be applied to non-monotonic alignment between signs and texts and benefit SLT?} 

Inspired by the recent success of joint CTC/Attention in translation tasks~\cite{yan-etal-2023-ctc}, we investigate the potential of this approach for SLT. This combination provides a complementary solution: during encoding, hierarchical encoding effectively manages alignment (including sign language length adjustment and reordering), while joint decoding reduces exposure/label bias in the attention decoder by leveraging CTC's conditionally independent alignment information. Our contributions are as follows:
\begin{itemize}
    \item We propose leveraging a joint CTC/Attention framework combined with transfer learning to address the unique challenges of SLT, particularly the modality gap between visual sign language and spoken text. Compared with speech and text translation, SLT requires handling more complex spatial-temporal representations and addressing non-monotonic alignments to grasp the linguistic feature embedded in sign language. 
 \item The proposed method achieves results comparable to state-of-the-art, outperforming the pure-attention baseline on widely adopted benchmarks, RWTH-PHOENIX-Weather 2014 T~\cite{camgoz2018neural} and CSL-Daily~\cite{zhou2021improving} (\S~\ref{sec:results}).
    \item We offer a promising direction for future research by exploring the potential of text CTC alignment for \textit{gloss-free} SLT.
    
\end{itemize}

\section{Background }

Both CTC~\cite{ctc} and the attentional encoder-decoder paradigm~\cite{vaswani2017attention} are designed to model Bayesian decision-making by selecting the output sequence $\hat{Y}$ that maximizes the posterior likelihood $P(Y|X)$ from all possible sequences $\mathcal{V}^\textrm{tgt}\ast$. Here, $X = \left \{ x_t\in \mathcal{S}^\textrm{src}|t=1,...,T \right \}$ and $Y = \left \{ y_l \in \mathcal{B}^\textrm{tgt}|l=1,...,L  \right \}$, where $\mathcal{S}^\textrm{src}$ and $\mathcal{B}^\textrm{tgt}$ denote the source and target language vocabularies, respectively, and $T$ and $L$ represent the lengths of the source and target sequences.
The differences between the CTC and attention frameworks, a typical method of it is transformer-based encoder-decoder model, are illustrated in the following sections.

\subsection{Comparison between CTC and Attention Framework}
\paragraph{Hard vs. Soft Alignment.} CTC marginalizes the likelihood over all possible input-to-alignment sequences $Z = \left \{z_t\in \mathcal{B}^\text{tgt}\cup \left \{ \textrm{blank}\right \}  | t = 1,...,T    \right \}  $, by introducing  the blank label~\cite{hannun2017sequence}. Each output unit $z_t$ (whether blank or non-blank) is mapped to a single input unit $x_t$ in a monotonic manner, resulting in hard alignment. In contrast, the attention framework allows a flexible input-to-output mapping via soft alignment, where an output unit $y_l$ maps multiple input units $x_{[...]}$ or vice versa.

\paragraph{Conditional Independence vs. Conditional Dependence.} CTC assumes conditional independence, meaning that each $z_t$ is independent of $z_{1:t-1}$ if it is already conditioned on $X$~\cite{nozaki2021relaxing}. While the attention framework models each output unit $y_l$ with conditional dependence on both the input $X$ and previous output units $y_{l:l-1}$. This conditional dependence introduces potential label biases~\cite{andor-etal-2016-globally}, where the model may over-rely on previous outputs during decoding.

\paragraph{Input-synchronous Emission vs. Autoregressive Generation.} CTC is an input-synchronous model, producing an output unit at each input time step. As a result, CTC cannot generate an output sequence longer than the input, which simplifies the process by eliminating the need for explicit end detection. Conversely, the decoder in the attention framework is autoregressive, generating outputs one by one until a stop token \texttt{<eos> }is emitted.

As shown in~\cite{yan-etal-2023-ctc}, joint modeling of CTC/Attention during encoding and decoding benefits translation. During encoding, the hard alignment provided by CTC produces stable representations, enabling the decoder to learn soft alignment patterns more effectively. In joint decoding, the conditionally independent likelihood of CTC helps mitigate label biases in the conditionally dependent framework of attention. Additionally, the end detection issue in autoregressive generation is alleviated by the interaction between input-synchronous emission and autoregressive generation. \textit{Will these advantages hold in SLT?}


\subsection{Sign Language Translation}
\label{subsection:slt}
SLT can be viewed as a MT problem, with the distinction that the source language is represented by spatial-temporal pixels rather than discrete tokens~\cite{chen2022simple}. Intuitively, the joint CTC/Attention approach is well-suited for SLT, as it integrates the robust alignment capabilities of CTC with the contextual understanding provided by attention, effectively addressing the unique challenges of translating sign language into spoken text.

Our study focuses on translating a sign language video into a spoken text in an end-to-end manner, taking into account the information bottleneck that gloss introduces in cascading SLT, where gloss serves as the intermediate supervision~\cite{yin2020better}. Many end-to-end SLT approaches~\cite{camgoz2020sign,zhang2023sltunet,ye-etal-2023-cross,tan-etal-2024-seda} model the conditional dependent likelihood to generate spoken text. In general, given a sign video sequence $\mathcal{V}=\left \{ v_1, v_2,..., v_{|\mathcal{V}|}  \right \} $ consisting of $|\mathcal{V}|$ frames, the SLT model is designed to predict the ground truth reference $\mathcal{T} = \left \{ t_1, t_2,..., t_{|\mathcal{T}|}  \right \}$ which contains $|\mathcal{T}|$ words. A gloss sequence $\mathcal{G} = \left \{ g_1, g_2,..., g_{|\mathcal{G}|}  \right \} $ of length $|\mathcal{G}|$ employed as an additional regularization term. The core of these SLT approaches is a shared encoder (see Figure~\ref{fig:model} left), trained with a joint loss to produce an encoded representation $h$, which is then used to predict both the gloss sequence $\mathcal{G}$ and spoken text $\mathcal{T} $. For each time step $l$, there are:
\begin{equation}
  \label{eq:ctc_ori}
  P_\text{CTC}(g_l|\mathcal{V}) = \text{CTC}(h_l),
\end{equation}
\begin{equation}
  \label{eq:Attn_ori}
  P_\text{Attn}(t_l|\mathcal{V},t_{1:l-1}) = \text{Decoder}(h,t_{1:l-1}),
\end{equation}
where \text{CTC}($\cdot$) refers to a projection onto the CTC output gloss $\mathcal{G}\cup \left \{ \textrm{blank} \right \} $ followed by softmax. Meanwhile, Decoder($\cdot$) represents autoregressive decoder layers followed by a projection to the decoder output and softmax. For a given training sample consisting of a sign video, gloss sequence, and text, $(\mathcal{V},\mathcal{G},\mathcal{T})$, the whole SLT model is optimized using the maximum likelihood estimation (MLE) loss for SLT, and the CTC loss for SLR. The total training loss is formulated as:
\begin{equation}
  \label{eq:s2t}
  \mathcal{L}(\mathcal{T,G}|\mathcal{V}) = \mathcal{L}_\text{MLE}(\mathcal{T}|\mathcal{V}) + \alpha\mathcal{L}_\text{CTC}(\mathcal{G}|\mathcal{V}),
\end{equation}
where hyperparameter $\alpha$ is utilized to balance the regularization effect of the CTC loss. 

\section{Model Architecture}
\begin{figure}[tb]
  \includegraphics[width=65mm]{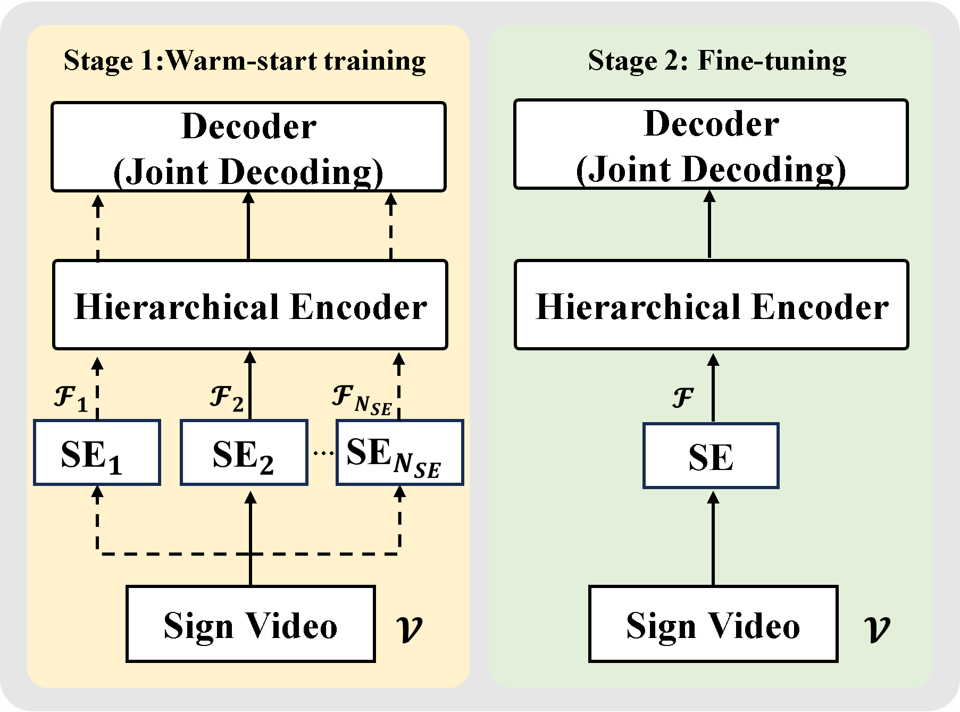}
  \centering
  \caption{Overview of the proposed method: Stage 1: warm-start training using multiple sign embeddings with a hierarchical encoder and joint decoding. Stage 2: fine-tuning the model with a single sign embedding.}
  \label{fig:overview}
\end{figure}

As shown in Figure~\ref{fig:overview}, our proposed method comprises three modules: sign embedding (SE), a hierarchical encoder conducting length adjustment and reordering, and a decoder that involves joint scoring during decoding. The training process is divided into two stages. Stage 1: Warm-start training, where the parent model is trained using sign videos with multiple SEs ($SE_1, SE_2,..., SE_{N_{SE}}$). Each SE extracts different representations ($\mathcal{F}_1, \mathcal{F}_2, ..., \mathcal{F}_{N_{SE}}$) from the sign video ($\mathcal{V}$), which are then passed through the hierarchical encoder and decoder. Stage 2: Fine-tuning the pre-trained model from Stage 1 with a single SE and child data pairs.
\subsection{Sign Embedding Module}
Since the sign video sequence is significantly longer than both the gloss and text sequences, \emph{i.e.} $|\mathcal{V}|\gg|\mathcal{G}| $ and $|\mathcal{V}|\gg|\mathcal{T}|$, we down-sample the sign sequences to align them with the shorter gloss and text sequences. Depending on the training stage, we utilize either multiple or a single pre-trained visual model to extract frame-level representations from a sign video. The number of SEs ($N_{SE}$) is determined by dataset and model availability (\S~\ref{sec: sign embeddings}). Following previous work, we use pre-trained SEs to extract sign representations from the video frames, which are then down-sampled through a linear layer. This process is formalized as:
\begin{equation}
    \mathcal{F} = \text{SignEmbedding}(\mathcal{V}),
    \label{se}
\end{equation}
where $\mathcal{F} = \left \{ f_1, f_2,..., f_{|\mathcal{F}|}  \right \} $ denotes sign representations after the pre-trained SEs and subsequent down-sampling. Note that the parameters of the pre-trained SEs remain frozen throughout both training stages.
\subsection{Hierarchical Encoding Module}
\begin{figure}[t]
  \includegraphics[width=\columnwidth]{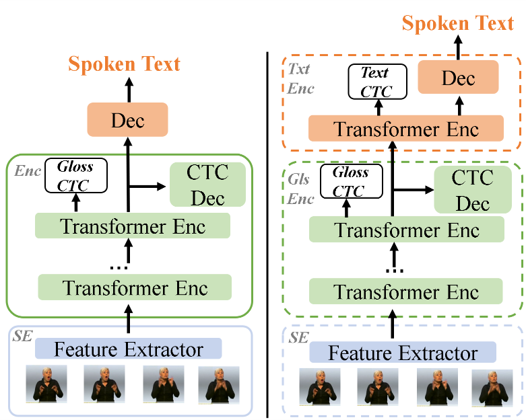}
  \centering
  \caption{The shared SLT encoders (left): sign representations are down-sampled by a shared encoder using gloss CTC; hierarchical SLT encoders (right): sign representations are first down-sampled by \textit{GlsEnc} using gloss CTC (\textit{GlsCTC}) and then reordered by \textit{TxtEnc} with text CTC (\textit{TxtCTC}).}
  \label{fig:model}
\end{figure}

Unlike the previous work on SLT adopting the shared encoder, we introduce a hierarchical encoder, which has been widely utilized in automatic speech recognition (ASR)~\cite{krishna2018hierarchical,higuchi2022hierarchical}. As shown in Figure~\ref{fig:model} (right), our hierarchical encoder consists of gloss-oriented encoder (\textit{GlsEnc}) and text-oriented encoder (\textit{TxtEnc}) layers responsible for sign representations length adjustment and reordering, respectively.

\paragraph{Gloss-oriented Encoder.}
The \textit{GlsEnc} is trained using the CTC criterion, which aligns intermediate encoder representations with gloss sequences. The representations from the intermediate recognition encoder layers are then passed to the CTC decoder to generate gloss sequences.
\paragraph{Text-oriented Encoder.}
Next, \textit{TxtEnc} reorders the representations using hierarchical encoder layers. Here, the CTC criterion aligns the final encoder representations with spoken text sequences. While CTC is generally restricted to monotonic alignments, the neural network architecture enables latent reordering to handle the non-monotonic nature of text alignment. Our proposed SLT hierarchical encoder consists of the following components:
\begin{equation}
h^{\textit{gls}} = \textit{Gls}\text{Enc}(f),
    \label{glsEnc}
\end{equation}
\begin{equation}
P_{\text{CTC}}(g|f) = \textit{Gls}\text{CTC}(h^{\textit{gls}}),
    \label{glsCTC} 
\end{equation}
\begin{equation}
h^{\textit{txt}} = \textit{Txt}\text{Enc}(h^{\textit{gls}}),
    \label{txtEnc}
\end{equation}
\begin{equation}
P_{\text{CTC}}(z_t^{\textit{txt}}|f) = \textit{Txt}\text{CTC}(h^{\textit{txt}}).
    \label{txtCTC}
\end{equation}
The \textit{Gls}Enc comprises $N_\textit{Gls}$ Transformer layers and a CTC loss calculation, while \textit{Txt}Enc consists of $N_\textit{Txt}$ Transformer layers and a CTC loss calculation. We newly introduce \textit{TxtCTC} training to facilitate effective reordering in the encoding process. Using a multi-task objective, both hierarchical encoder and attentional decoder are jointly optimized:
\begin{equation}
\mathcal{L}_{\text{SLT}} = \lambda_1\mathcal{L}_{\textit{GlsCTC}}+\lambda_2\mathcal{L}_{\textit{TxtCTC}}+\lambda_3\mathcal{L}_{\textit{MLE}},
    \label{Total_loss}
\end{equation}
where $\mathcal{L}_{\textit{MLE}}$ is a MLE loss for SLT, and $\lambda$s decide the effects of gloss-oriented CTC, text-oriented CTC, and decoder MLE losses.
\subsection{Joint Decoding Module}

Our joint decoding adopts the \textit{out-synchronous decoding} (\emph{i.e.} attention plays a leading role) from~\cite{yan-etal-2023-ctc} for its superior performance. However, ours differs in candidate selection: we first use the attention mechanism to propose candidates, then apply \textit{Txt}CTC constraints, offering greater flexibility and reducing label bias. 

As shown in Algorithm~\ref{alg: jointdecoding}, the joint decoding process has four steps: candidate selection, hypotheses expansion, joint scoring, and end detection. The algorithm generates candidates using the attentional decoder at each step \textit{l}. A pre-defined beam size $N_\text{beam}$ is adopted to select the top candidates (cand), starting with the partial hypotheses (hyp) from the previous step. Each hypothesis is extended by concatenating ($\oplus$) it with new candidates, adjusted by a dynamically computed length penalty $\beta$. The candidates are then scored using a joint mechanism $P_\text{beam}$ combining normalized \textit{Txt}CTC scores (${\hat{\alpha}}_{\text{CTC}}$) and Attention scores (${\hat{\alpha}}_{\text{Attn}}$), where \textit{Txt}CTC ($\alpha_\text{CTC}$) and Attention scores ($\alpha_\text{Attn}$) represent their respective likelihoods. This joint scoring guides the selection of promising hypotheses. Updated partial hypotheses are stored in newHyp, while completed hypotheses ending with <eos> or reaching $L_{\textit{max}}$ are stored in finHyp.

\begin{algorithm}[t]
    \caption{Joint Beam Search Decoding: The attentional decoder first proposes candidates for expanding hypotheses, ensuring all have \textit{l}-length at step \textit{l}.}
    \small
    \label{alg: jointdecoding}
    \begin{algorithmic}[1]
        \Procedure{Joint Beam Search Decoding}{hyp, $h^\textit{txt}$,  $L_{\textit{max}}$, $N_{\text{beam}}$}
        \State newHyp=\{\}; finHyp=\{\}
        \For{$t_{1:l-1} \in$ hyp}
            \State cand = top-k($P_\text{attn}(t|h^\textit{txt}, t_{1:l-1})$, k=$N_{\text{beam}}$)
            \State $\beta = \text{Length\_Penalty}(l+1)$
            \State cand = cand + $\beta$ 
            \For{$c\in$ \text{cand}}                    
                \State $t_{1:l}=t_{1:l-1}\oplus c$ 
                \State $\alpha_\text{CTC} = \text{TxtCTCScore}(t_{1:l})$
                \State $\alpha_\text{Attn} = \text{AttnScore}(t_{1:l})$
                \State $P_{\text{Beam}}$ = $\hat{\alpha}_\text{CTC}+\hat{\alpha}_\text{Attn}$ 
                \If{(c \text{is <eos>) or ($l$ is $L_{\textit{max}}$)}} 
                \State finHyp $\left [  t_{1:l}\right ] $ =$P_{\text{Beam}}$($\cdot $)
                \Else  
                \State newHyp$\left [  t_{1:l}\right ] $ = $P_{\text{Beam}}$($\cdot $)
                \EndIf
            \EndFor                    
        \EndFor
        \State\Return newHyp, finHyp
        \EndProcedure
    \end{algorithmic}
\end{algorithm}
\subsection{Transfer Learning}
 One of the significant obstacles hindering the advancement of SLT is data scarcity~\cite{moryossef2021data, zhu-etal-2023-neural}. To address this, many SLT studies mitigate the low-resource issue by transfer learning~\cite{chen2022simple,zhang2023sltunet}. Transfer learning is widely used in low-resource scenarios to enhance model performance by leveraging knowledge from related tasks.

\paragraph{Warm-start Training.}
In transfer learning~\cite{10.1162/coli_a_00446}, a parent model is pre-trained on a large corpus, and its parameters are used to initialize the child model on a smaller, target-specific corpus. Warm-start training goes a step further by incorporating child language pairs into the parent model’s pre-training. After pre-training, the model is fine-tuned exclusively on the child language pairs. This approach benefits from the high likelihood of word or character overlap between those languages, improving the efficiency of transfer learning~\cite{neubig-hu-2018-rapid}. For this reason, we use parallel datasets to paraphrase spoken text by introducing normalization, lemmatization, and back translation. We adopt multiple pre-trained SEs for sign videos to extract varied sign representations from the same video. 
\paragraph{Fine-tuning.}
The parent model is trained using sign representations from multiple SEs and combined spoken texts, while the child model is fine-tuned with a single SE and the original corpus. 


\begin{table*}[tbh]
\centering
\scalebox{0.75}{
\begin{tabular}{lrrr|rrr||rrr|rrrr}
\toprule
\textbf{} & \multicolumn{6}{c||}{\textbf{PHOENIX14T}} & \multicolumn{6}{c}{\textbf{CSL-Daily}} \\

\textbf{} & \multicolumn{3}{c|}{\textbf{Text}}& \multicolumn{3}{c||}{\textbf{Augmented text}} & \multicolumn{3}{c|}{\textbf{Text}} & \multicolumn{3}{c}{\textbf{Augmented text}} \\ \hline
\textbf{} & \textbf{Train} & \textbf{Dev} & \textbf{Test} & \textbf{Train} & \textbf{Dev} & \textbf{Test} & \textbf{Train} & \textbf{Dev} & \textbf{Test} & \textbf{Train} & \textbf{Dev} & \textbf{Test}\\ \hline
\textbf{Sentences} & 7,096 & 519 & 642 & 21,288 &519& 642 & 184,401 & 1,077 & 1,176 & 36,802 & 1,077 & 1,176\\ \hline
\textbf{Vocab.} & 2,887 & 951 & 1,001 & 3,770&  951 & 1,001 & 2,343 & 1,358 & 1,358 &  2,491 & 1,358 & 1,358  \\ \hline
\textbf{\#Words} & 99,081 & 6,820 & 7,816 & 308,198 & 6,820 & 7,816& 291,048 &  17,304 &  19,288 &  362,778 & 17,304 &  19,288\\ \hline
\textbf{\#OOVs} & -- & 57 & 60 & -- & 55 & 58& -- & 64 & 69 & -- & 43 & 47 \\ 
\bottomrule
\end{tabular}}
\caption{Statistics of the augmented corpora. "Sentences": the total number of sentences; "Vocab.": the number of spoken words (for Chinese, we count characters); "\#Word": the total number of words; and "\#OOV": the out-of-vocabulary words that appear in the development and test sets but are absent in the training set.}
\label{tab:corpora_stats}
\end{table*}

\begin{table*}   
\centering
\scalebox{0.8}{
\begin{tabular}{llcc}
\toprule
\multirow{2}{*}{\textbf{ID}} & \multirow{2}{*}{\textbf{\textit{Systems} }}& \textbf{PHOENIX14T} &\textbf{CSL-Daily} \\
&&B@4 $\uparrow$&B@4 $\uparrow$\\
\midrule
1 & Baseline & 20.96& -- \\

\midrule
\textit{Explore Sign Embeddings} & & &\\
2& 1 + Replace sign embeddings to SMKD model &24.32&15.54 \\

2.1& 1 + Replace sign embeddings to Corrnet model &24.64&13.02 \\
2.2& 1 + Combine all sign embeddings &24.38 & 13.77 \\

\midrule
\textit{Explore CTC Capacity}&&&\\
\small\textit{ Hierachical encoding}&&&\\
3& 2 + \textit{GlsCTC}  &24.87 &16.61 \\
3.1 & 1 + \textit{GlsCTC}&22.40&--\\
3.2& 2.1 + \textit{GlsCTC}&24.70&16.43\\
3.3& 2.2 + \textit{GlsCTC}&23.26 & 14.38\\
4 & 2  + \textit{TxtCTC}  &25.00 &15.36 \\
5& 2 + \textit{GlsCTC} and \textit{TxtCTC}  &25.84 &18.27 \\

\small\textit{ Joint decoding}&&&\\
6& 4 + Joint decoding  &25.63 &15.79 \\
7& 5 + Joint decoding  &26.37 &19.17 \\
\midrule
\textit{Explore Transfer Learning}&&&\\
8 & 3.2 + Warm start training & 24.24& 17.38\\  
8.1 & 8 + Fine-tuning & 25.26 & 18.01 \\
9 & 7 + Combine all sign embeddings + warm start training &27.07& 20.29\\  
\midrule
\textit{Joint CTC/Attention}&&&\\
10 & 9 + Fine-tuning &\textbf{ 27.93} & \textbf{22.04} \\
\bottomrule
\end{tabular}}
\caption{Ablation study of the proposed methods on dev sets. B@4: tokenized 4-gram BLEU.}
\label{tab:results}
\end{table*}

\section{Experimental Setup}
\paragraph{Dataset.} We evaluate the effectiveness of our proposed method on two popular benchmarks for SLT, RWTH-PHOENIX-Weather 2014 T (PHOENIX14T)~\cite{camgoz2018neural} and CSL-Daily~\cite{zhou2021improving}. PHOENIX14T focuses on German sign language, sourced from weather broadcasting and CSL-Daily focuses on Chinese sign language, recorded in an ad hoc environment. Both datasets provide triplet samples: sign language video, a sentence-level gloss annotation, and corresponding spoken text ($\mathcal{V}, \mathcal{G}, \mathcal{T}$).
\paragraph{Text Preprocessing.} Following~\citet{zhu-etal-2023-neural}, we perform lemmatization and alphabet normalization on PHOENIX14T. Additionally, we apply back-translation using a text-to-gloss translation model to generate new source sentences from the glosses on the target side.  For CSL-Daily, we follow a similar procedure, using a Zh-En model~\cite{backtranslation} to generate parallel spoken text. The combined data is then used in warm-start training. Table~\ref{tab:corpora_stats} shows the statistics of the combined corpora.

\paragraph{Evaluation Metrics.} We report results on SLT tasks using tokenized BLEU with n-grams from 1 to 4 (B@1-B@4)~\cite{papineni2002bleu} and Rouge-L F1 (ROUGE)~\cite{lin2004rouge}. We employ word error rate (WER) to evaluate SLR performance.

\paragraph{Sign Embedding.} Following recent studies~\cite{camgoz2020sign, zhou2021improving,zhang2023sltunet} that use pre-trained models to extract sign video representations, we adopt multiple pre-trained visual models as SEs for warm-start training. For PHOENIX14T, we utilize pre-trained CNN layers from the CNN+LSTM+HMM setup~\cite{CNNpretrain}, the pre-trained SMKD model\footnote{\url{https://github.com/VIPL-SLP/VAC_CSLR}}~\cite{SMKD}, and the pre-trained Corrnet model\footnote{\url{https://github.com/hulianyuyy/CorrNet}}~\cite{corrnet}. For CSL-Daily, we adopt the pre-trained SMKD model and Corrnet to extract sign representations. 
The parameters of these pre-trained models are frozen throughout the training process. The child model utilizes a single SE, selected based on initial performance (\S~\ref{sec: sign embeddings}).

\paragraph{Modeling.} We compare our joint CTC/Attention
model with purely attentional encoder-decoder baseline, Sign Language Transformer~\cite{camgoz2020sign}. The proposed and baseline models are trained separately with the same hyperparameters. Full details of the model sizes and hyperparameters are provided in the Appendix Table~\ref{tab: Hayperparameters} and~\ref{tab: Hayperparameters of decoding}. Unless otherwise mentioned, we maintain these setups for the ensuing experiments.

\begin{table*}[tbh]
\centering
\begin{threeparttable}
\scalebox{0.85}{
\begin{tabular}{lcc|ccccc}
\toprule
\multirow{2}{*}{\textbf{Methods}}&\multicolumn{2}{c}{\textbf{DEV}} &\multicolumn{5}{c}{\textbf{TEST}} \\

&B@4&ROUGE&B@1&B@2&B@3&B@4&ROUGE\\
\midrule%
\textbf{\textit{Cascading}}&&&&\\
STMC-Transformer~\cite{yin2020better}& 24.68& 48.70& 50.63&38.36& 30.58&25.40&48.78\\
ConSLT~\cite{conslt}&24.27& 47.52&--&--&--&25.48& 47.65\\
\midrule
\textbf{\textit{End-to-end (Pure Attention)}}&&&&&&\\
Joint-SLRT~\cite{camgoz2020sign}&22.38&--&46.61&33.73&26.19&21.32&-- \\
Sign Back Translation~\cite{zhou2021improving}& 24.45&50.29&50.80&37.75&29.87&24.34&49.54\\
STMC-T~\cite{STMC-T}&24.09&48.24&46.98&36.09&28.70&23.65&46.65 \\
PET~\cite{jin2022priorPET}&--&--&49.54&37.19&29.30&24.02&49.97\\
VL-Transfer~\cite{chen2022simple}&27.61&53.10&\textbf{53.97}&41.75&33.84&28.39&52.65 \\
SLTUNet~\cite{zhang2023sltunet}&27.87& 52.23 &52.92&41.76&\textbf{33.99}&\textbf{28.47}&52.11 \\
XmDA~\cite{ye-etal-2023-cross}&25.86&52.42 &--&--&--&25.36& 49.87 \\

\midrule

\textbf{\textit{Ours (Joint CTC/Attention)}}&\textbf{27.93}& \textbf{53.53}&53.95&\textbf{41.81}&33.86& 28.42&\textbf{53.34}\\

\bottomrule
\end{tabular}} 

\end{threeparttable}
\caption{ Results of different systems on PHOENIX14T. }
\label{table: End-to-end SLU performance on PHOENIX14T dataset}
\end{table*}
\section{Results and Analyses}
\label{sec:results}
We validate our proposed method on two popular benchmarks, PHOENIX14T and CSL-Daily, comparing it with both \textit{cascading} and \textit{end-to-end} state-of-the-art methods.\footnote{The distinction between cascading and end-to-end is in \S~\ref{subsection:slt}. Note that scores from previous papers are reported.}

\subsection{Experimental Results}
Our proposed method outperforms pure-attention encoder-decoder models on PHOENIX14T (see Table~\ref{table: End-to-end SLU performance on PHOENIX14T dataset}) and achieves results comparable to the state-of-the-art on CSL-Daily, as shown in Table~\ref{table: End-to-end SLU performance on CSL-Daily dataset}. We conducted a series of ablation studies on their development sets, summarized in Table~\ref{tab:results}.  On PHOENIX14T, it surpasses the baseline by 6.97 BLEU scores (Table~\ref{tab:results}: 1$\rightarrow$10), demonstrating its effectiveness, particularly in handling non-monotonic alignment in SLT. Similarly, on CSL-Daily, consistent improvements with a gain of 6.50 BLEU scores are observed (Table~\ref{tab:results}: 2$\rightarrow$10). 

 
\begin{table*}[ht]
\centering
\begin{threeparttable}
\scalebox{0.85}{
\begin{tabular}{lcc|ccccc}
\toprule
\multirow{2}{*}{\textbf{Methods}}&\multicolumn{2}{c}{\textbf{DEV}} &\multicolumn{5}{c}{\textbf{TEST}} \\

&B@4&ROUGE&B@1&B@2&B@3&B@4&ROUGE\\
\midrule%
\textbf{\textit{Cascading}}&&&&\\
ConSLT~\cite{conslt}&14.80&  41.46&--&--&--&14.53& 40.98\\
\midrule
\textbf{\textit{End-to-end (Pure Attention)}}&&&&&&\\
Joint-SLRT~\cite{zhou2021improving}&11.88&27.06&37.38&24.36&16.55&11.79&36.74 \\
Sign Back Translation~\cite{zhou2021improving}&20.80&49.49&51.42&37.26&27.76& 21.34&49.31\\
XmDA~\cite{ye-etal-2023-cross}&21.69&49.36 &--&--&--&21.58& 49.34 \\

\midrule

\textbf{\textit{Ours (Joint CTC/Attention)}}&\textbf{22.04}&\textbf{51.62}&\textbf{52.37}&\textbf{38.22}&\textbf{29.72}&\textbf{22.47}&\textbf{51.87}\\ 

\bottomrule
\end{tabular}} 
\end{threeparttable}
\caption{ Results of different systems on CSL-Daily.}
\label{table: End-to-end SLU performance on CSL-Daily dataset}
\end{table*}
\subsection{Exploring Sign Embeddings}
\label{sec: sign embeddings}
\begin{table}[tbh]
    \centering
    \scalebox{0.8}{
    \begin{tabular}{l|cc|cc}
    \toprule
    \multirow{2}{*}{\textbf{Sign embedding (ID)} }&\multicolumn{2}{c|}{\textbf{PHOENIX14T} } &\multicolumn{2}{c}{  \textbf{CSL-Daily}}\\
    & WER$\downarrow$&B@4$\uparrow$& WER$\downarrow$&B@4$\uparrow$\\
    \midrule
   Pre-trained CNN (3.1)&48.31&22.40& -- & -- \\
    SMKD (3)&21.88&\textbf{24.87}&28.48&\textbf{16.61}\\
    Corrnet (3.2)&\textbf{21.43}&24.70&30.54&16.43\\
    Combining All (3.3)&23.01&23.26&\textbf{28.16}	&14.38\\
    \bottomrule
    \end{tabular}}
    \caption{SLR performance of different sign embeddings.}
    \label{tab:Ablation on sign embedding}
\end{table}
As indicated by~\citet{zhang2023sltunet}, high-quality SE benefits SLT, we explored the impact of different SEs on both SLR and SLT by replacing the baseline SE. Among all SEs, the pre-trained SMKD model shows superior performance in SLT on both datasets. Therefore, we used the SMKD model as the default SE in further analyses unless stated otherwise. We also examined SLR performances of all SEs. Interestingly, while combining all SEs slightly improves SLR performance on CSL-Daily, it underperforms in SLT compared to single embeddings such as SMKD or Corrnet (see Table~\ref{tab:Ablation on sign embedding}). Since both SMKD and Corrnet are pre-trained using gloss annotations, their sign representations may share overlaps. In addition, glosses introduce information loss, limiting the benefit of combining these embeddings. However, as further discussed in \S~\ref{sec:transfer learning}, transfer learning with fine-tuning can mitigate this issue and enhance performance. 

\subsection{Exploring CTC Capacity}
\textit{TxtCTC} plays a crucial role in both hierarchical encoding and joint decoding. It serves as a reordering tool for sign representations during encoding and provides the CTC score during decoding. To fully understand how each CTC component functions and verify whether they achieve the intended goals of length adjustment and reordering, we examined the respective contributions of \textit{GlsCTC} and \textit{TxtCTC}.
First, we examined how both CTCs contribute to the model when only attention is used in decoding (Table~\ref{tab:results}: explore hierachical encoding). We then explored the contribution of \textit{TxtCTC} in joint decoding by performing joint decoding without hierarchical encoding; in this case, only \textit{TxtCTC} was involved in the whole process. Finally, we evaluated the performance of hierarchical encoding together with joint decoding.

The \textit{TxtCTC} on its own appears to contribute more to PHOENIX14T than it does to CSL-Daily (Table~\ref{tab:results}: 3 and 4), indicating that reordering is more critical for German sign language. Hierarchical encoding yields more quality gains on the CSL-Daily dataset (Table~\ref{tab:results}: 2$\rightarrow$5). Joint decoding with only \textit{TxtCTC} achieves 25.63 BLEU and 15.79 BLEU on PHOENIX14T and CSL-Daily, respectively. This demonstrates the reordering capacity of CTC, especially when combined with attention mechanism for decoding, benefits SLT. Since no gloss is involved during training in ID 6 of Table~\ref{tab:results}, joint decoding with \textit{TxtCTC} provides an alternative towards \textit{gloss-free} SLT, which often struggles with non-monotonic alignments (\emph{e.g.,}~\citet{gong2024llms} present their \textit{gloss-free} SLT achieves BLEU score of 25.25 and 12.23 on development sets of PHOENIX14T and CSL-Daily). The combination of hierarchical encoding and joint decoding boosts SLT performance by 0.53 BLEU on PHOENIX14T and 0.90 on CSL-Daily (Table~\ref{tab:results}: 5$\rightarrow$7). 
%



\subsection{Exploring Transfer Learning}
\label{sec:transfer learning}
In \S~\ref{sec: sign embeddings}, we demonstrated that simply combining all pre-trained SEs deteriorates SLT. We further investigated how this combination of SEs performs when paired with augmented spoken texts in transfer learning. For PHOENIX14T, we paired sign representations from pre-trained CNNs, Corrnet, and SMKD with pre-processed, back-translated, and original spoken texts, respectively. For CSL-Daily, we paired sign representations from Corrnet and SMKD with back-translated and original texts. As shown in Table~\ref{tab:results}, warm-start training mitigates the limitation of combining all SEs (+ 0.98 BLEU on PHEONIX14T, + 3.00 BLEU on CSL-Daily, Table~\ref{tab:results}: 3.3$\rightarrow$8). While warm-start training enhances SLT by sharing mixed parameters, it lacks task-specific characteristics. To address this, we fine-tuned using SMKD sign representations and original spoken text. This further improves performance (Table~\ref{tab:results}: 8$\rightarrow$8.1 and 9$\rightarrow$10). 

Besides, the case study of translation outputs is available in Appendix Table~\ref{tab:case study}.


\section{Related Work}
The exploration of CTC's latent alignment capacity has been widely conducted in speech translation (ST) and machine translation (MT)~\cite{haviv-etal-2021-latent}. In ST, several studies have explored CTC’s potential to align and reorder spoken inputs with their target language translations. \citet{chuang-etal-2021-investigating} found that CTC enhances the reordering behavior of non-autoregressive ST. Additionally, \citet{zhang2022revisiting} demonstrated the effectiveness of CTC in ST by jointly training CTC and attention mechanisms without relying on ASR transcriptions. In MT, CTC has gained attention for both non-autoregressive~\cite{saharia-etal-2020-non} and autoregressive paradigms~\cite{yan-etal-2023-ctc}. 

Despite CTC's reordering potential, research in SLT has underexplored this capability. In early SLT models, CTC was adopted primarily for its ability to align glosses with unsegmented sign video. \citet{camgoz2020sign} pioneered SLT by jointly training CTC and attention mechanisms with glosses and corresponding spoken texts. Several subsequent works~\cite{zhou2021improving,chen2022simple,zhang2023sltunet,ye-etal-2023-cross, tan-etal-2024-seda} followed this paradigm, incorporating CTC to localize glosses and introduce techniques such as transfer learning and data augmentation to advance SLT. However, in these models, CTC's role was limited to monotonic alignment, with little focus on its ability to reorder sequences. This approach, while functional, led to significant limitations in SLT. As noted by \citet{yin-etal-2021-including}, relying on glosses for SLT introduces an information bottleneck as the multi-dimensional sign video is converted into a linear gloss sequence. Additionally, glosses are linguistic tools not typically used by deaf and hard-of-hearing individuals for daily communication \cite{muller-etal-2023-considerations}, further limiting their practical utility in SLT.

Given the limitations of glosses and the resource-intensive nature of gloss annotation~\cite{uthus2024youtube}, recent research has shifted towards \textit{gloss-free} SLT, which aims to translate sign video into text without gloss. This shift introduces challenges in aligning sign video with textual sequence. \citet{wong2024sign2gpt} addressed this with a pseudo-gloss pretraining strategy in their Sign2GPT model for flexible alignment. Similarly, \citet{gong2024llms} employed a Vector-Quantized visual sign module with a Codebook Reconstruction and Alignment module. Although \textit{gloss-free} SLT has made significant progress, these approaches have moved away from using CTC, likely due to the assumption that CTC mainly supports monotonic alignment without fully exploring its reordering potential. 

\section{Conclusion}

Unlike prior SLT approaches, this work revisits the reordering capability of CTC in the context of sign language translation (SLT). We propose a Joint CTC/Attention framework combined with transfer learning, leveraging CTC's alignment and reordering capabilities for more effective SLT. Specifically, we introduce hierarchical encoding to manage length adjustment and reordering through the \textit{GlsCTC} and \textit{TxtCTC }training criteria, effectively addressing non-monotonic mappings in SLT. The joint decoding facilitates dynamic interaction between CTC and attention mechanisms. Additionally, transfer learning narrows the modality gap between vision (sign language) and language (spoken text). Our proposed method achieves translation results comparable to state-of-the-art. Moreover, we demonstrate the strong potential of \textit{TxtCTC} in joint decoding without gloss regularization, providing an alternative for \textit{gloss-free} SLT to tackle non-monotonic alignments between sign language and spoken text.


\section{Limitations}

The limitations of our work lie in two points.  First, data availability: beyond the inherent scarcity in sign language research, our approach relies on gloss and text annotations, available in only a few datasets. This limits the scalability and advancement of data-driven SLT models. Second, although our framework shows potential for \textit{gloss-free} SLT, the current pre-trained sign embeddings still rely on gloss supervision, potentially introducing an information bottleneck, as discussed in \S~\ref{sec: sign embeddings}. In future work, we aim to explore sign representation strategies towards \textit{gloss-free} SLT.

\section*{Acknowledgments}
This work was partially supported by JST CREST JPMJCR19K1.
The authors thank Yui Sudo for the insightful discussion and the anonymous reviewers for their valuable feedback.

\bibliography{custom}

\appendix
\section{Appendix}
\begin{table}[tbh]
    \centering
    \scalebox{0.8}{
    \begin{tabular}{lll}
        \hline
    \textbf{Hyperparameter}   & \textbf{PHOENIX14T} &\textbf{ CSL-Daily} \\
    \hline
    $N_\textit{SE}$   & 3 & 2\\
    $N_\textit{Gls}$   & 5 & 5\\
    $N_\textit{Txt}$   & 1 & 1\\
    $N_\text{Dec}$     & 6 & 6\\
    $\lambda_1$ &1.0&5.0\\
    $\lambda_2$ &1.0&2.0\\
    $\lambda_3$ &1.0&1.0\\
    $L_{\textit{max}}$& 30 &50\\
    Attention heads    &4  & 4\\
    Hidden size        & 256 & 256\\
    Feed-forward dimension& 4096&4096\\
    Activation function& ReLU & Softsign\\
    Learning rate & $1\cdot 10^{-3}$&$1\cdot 10^{-3}$\\
    Adam$\beta$ & (0.9,0.998)&(0.9,0.998)\\
    Initializer&Xavier&Xavier\\
    Init gain & 0.5 & 0.5\\    
    Dropout rate& 0.3 & 0.3\\
    Label smoothing    & 0.1  & 0.1\\
    Batch size & 64 & 128\\
    \#params& 28.95M &28.95M \\
    \hline
    \end{tabular}}
    \caption{Hyerparameters and trainable papramenters \#params of our SLT models.}
    \label{tab: Hayperparameters}
\end{table}
\begin{table}[tbh]
    \centering
   
    \scalebox{0.7}{
    \begin{tabular}{lll}
        \hline
    \textbf{Decoding Type} &\textbf{Hyperparameter}   & \textbf{ Value} \\
    \hline
    \multirow{2}{*}{Pure attention}   & Penalty & 0.6\\
   & Beam size &  [1, 2, 3, 4, 5, 6, 7, 8, 9, 10]\\
       \hline
    
    \multirow{3}{*}{Joint decoding}  & Penalty & 0.6\\
      & Beam size &  [1, 2, 3, 4, 5, 6, 7, 8, 9, 10]\\
     & CTC score weight & 0.3\\
    
    \hline
    \end{tabular}}
     \caption{Decoding hyperparameters for pure attention and joint decoding methods.}
    \label{tab: Hayperparameters of decoding}
\end{table}
\begin{table*}
  \centering
  
   \small \begin{tabular}{ll}
    \hline

    \textbf{\textit{Systems}} & \textbf{Translation Output} \\
    \hline

    \textit{Examples from PHOENIX14T}&\\
    \multirow{2}{*}{Reference}   &  dort morgen neunzehn im breisgau bis siebenundzwanzig grad\\
    &\textit{(there tomorrow nineteen in Breisgau up to twenty-seven degrees)}\\

   \multirow{2}{*}{Baseline}      & morgen neunzehn grad im breisgau bis siebenundzwanzig grad im breisgau        \\
   &\textit{( tomorrow nineteen degrees in Breisgau up to twenty-seven degrees in Breisgau)}\\
  
    \multirow{2}{*}{Joint CTC/Attention}    & dort morgen bis neunzehn im breisgau bis siebenundzwanzig grad\\
    &\textit{(there tomorrow up to nineteen in Breisgau up to twenty-seven degrees)}\\
        \hline\\
        \multirow{2}{*}{Reference}   &  morgen regnet es vor allem noch in der südosthälfte  \\
    &\textit{(tomorrow it will still be raining, especially in the southeast)}\\
   \multirow{2}{*}{Baseline}      & in the southeast it rains occasionally      \\
  & \textit{(in the southeast it rains occasionally)}\\
        \multirow{2}{*}{Joint CTC/Attention}    & in der südosthälfte regnet es ergiebig  \\
    &\textit{(heavy rain in the southeast)}\\
\hline\\
\multirow{2}{*}{Reference}   &  in der nacht ist es teils wolkig teils klar stellenweise gibt es schauer oder gewitter\\
    &\textit{(during the night it will be partly cloudy partly clear with showers or thunderstorms in places)}\\

\multirow{2}{*}{Baseline}      & in der nacht ist es teils wolkig teils klar teils wolkig mit schauern und gewittern         \\
   &\textit{(at night it will be partly cloudy partly clear partly cloudy with showers and thunderstorms)}\\
   \multirow{2}{*}{Joint CTC/Attention}    & in der nacht teils wolkig teils klar örtlich schauer oder gewitter  \\
    &\textit{(at night partly cloudy partly clear local showers or thunderstorms
)}\\
 \hline
   \textit{Examples from CSL-Daily}&\\
   \multirow{2}{*}{Reference}   &  \begin{CJK*}{UTF8}{gbsn}蛋 糕 太 甜 ， 吃 多 了 对 身 体 不 好 。 \end{CJK*}\\  
    &\textit{(Cake is too sweet and too much of it is not good for your health.)}\\
    \multirow{2}{*}{Baseline}   &  \begin{CJK*}{UTF8}{gbsn}蛋 糕 太 甜 ， 很 安 静。 \end{CJK*}\\  
    &\textit{(The cake is too sweet and it's quiet.)}\\
    \multirow{2}{*}{Joint CTC/Attention}   &  \begin{CJK*}{UTF8}{gbsn}蛋 糕 太 甜 了， 身 体 不 健 康 \end{CJK*}\\  
    &\textit{(The cake is too sweet. It's not healthy.)}\\
\hline    
\\
   \multirow{2}{*}{Reference}   &  \begin{CJK*}{UTF8}{gbsn}我 自 愿 加 入 组 织 。 \end{CJK*}\\  
    &\textit{(I volunteer to join the organization.)}\\
    \multirow{2}{*}{Baseline}   &  \begin{CJK*}{UTF8}{gbsn}我 自 愿 参 加 这 项 危 险 的 任 务 。 \end{CJK*}\\  
    &\textit{(I volunteer for this dangerous mission.)}\\
    \multirow{2}{*}{Joint CTC/Attention}   &  \begin{CJK*}{UTF8}{gbsn}我 自 愿 参 加 的 组 织 。\end{CJK*}\\  
    &\textit{(I volunteer for the organization.)}\\
\hline  \\
 \multirow{2}{*}{Reference}   &  \begin{CJK*}{UTF8}{gbsn}他 将 来 想 成 为 科 学 家 。 \end{CJK*}\\  
    &\textit{(He wants to be a scientist in the future.)}\\
    \multirow{2}{*}{Baseline}   &  \begin{CJK*}{UTF8}{gbsn}他 想 做 点 兼 职 工 作 。\end{CJK*}\\  
    &\textit{(He wants to do some part-time work.)}\\
    \multirow{2}{*}{Joint CTC/Attention}   &  \begin{CJK*}{UTF8}{gbsn}他 以 后 想 做 科 学 校 。\end{CJK*}\\  
    &\textit{(He wants to be a science school in the future.)}\\
\hline

  \end{tabular}
  \caption{Case study of translation outputs on PHOENIX14T and CSL-Daily. Examples are from the development sets.
Sentences in bracket are our English approximation translations. Our Joint CTC/Attention with transfer learning generally outperforms the baseline by providing translations that are more detailed, contextually accurate, and faithful to the original semantics of the source text. }
\label{tab:case study}
\end{table*}

\end{document}